\documentclass{article} 
\usepackage{iclr2027_conference,times}

\usepackage{amsmath,amsfonts,bm}

\def\eqref#1{equation~\ref{#1}}

\def\1{\bm{1}}

\DeclareMathAlphabet{\mathsfit}{\encodingdefault}{\sfdefault}{m}{sl}
\SetMathAlphabet{\mathsfit}{bold}{\encodingdefault}{\sfdefault}{bx}{n}

\usepackage{hyperref}
\usepackage{url}
\usepackage{booktabs}
\usepackage{multirow}
\usepackage{graphicx}
\usepackage{xcolor}

\title{Enhancing Long-Video VLM Embeddings with Query-Aware Streaming Latent Reasoning}

\author{%
Haozhe Chi$^{1}$, Song Jin$^{2}$, Yang Jin$^{1}$, Yadong Mu$^{1}$\\
$^{1}$Peking University\\
$^{2}$Renmin University of China
}

\iclrfinalcopy 
\begin{document}

\maketitle
\lhead{}

\begin{abstract}
Long-video embedding requires capturing sparse query-relevant evidence under a limited visual-token budget. Uniform sampling can miss brief events in videos spanning minutes or hours, whereas encoding more frames in a single context increases memory and computation. We introduce \textbf{Query-Aware Streaming Latent Reasoning} (QASLR), a post-training framework that accumulates evidence across clips while keeping the embedding size fixed. QASLR selects a bounded set of frames, restores their temporal order, and processes them clip by clip with a vision-language backbone. A compact set of persistent think tokens cross-attends to each clip's features, while an embed token reads out a normalized representation after every update. This design integrates evidence across multiple backbone calls without requiring all selected frames to share a single context. Training combines final contrastive learning, step-wise contrastive supervision, and final-embedding self-distillation. Intermediate supervision trains partial-video readouts for retrieval, while self-distillation regularizes them toward the final representation. Query-aware selection produces query-conditioned representations for candidate-set scoring and reranking, whereas query-independent selection enables reusable corpus indexing. Under the full training recipe, HourVideo retrieval Hit@1 increases from 54.2 to 70.7 and from 57.4 to 72.8 for 2B and 8B Qwen3-VL-Embedding backbones, respectively. Gains extend to the evaluated moment-retrieval and video-QA tasks, and the streaming head transfers to a second Qwen-family embedding backbone. These results support streaming latent aggregation as an effective approach to integrating long-video evidence into fixed-dimensional representations.
\end{abstract}

\section{Introduction}
Multimodal embedding models provide a common representation interface for retrieving, organizing, and comparing text, images, and videos~\citep{radford2021clip,jia2021align,girdhar2023imagebind,meng2025vlm2vecv2}. Their fixed-dimensional outputs are attractive for large collections because candidate representations can be indexed once and searched efficiently~\citep{johnson2019faiss,karpukhin2020dpr}. Extending this interface to long videos is difficult, however, because the visual input grows with duration whereas the model's visual-token budget remains bounded~\citep{zhang2025longva,shen2025longvu}. A video may span tens of minutes, while the evidence that determines its relevance to a query may occupy only a few seconds~\citep{chandrasegaran2024hourvideo,wu2024longvideobench}. Consequently, the quality of a long-video embedding depends not only on semantic alignment, but also on whether the model observes the relevant evidence and retains it while processing the rest of the sequence.

Existing pipelines usually address the input constraint by sampling a fixed number of frames. Uniform sampling provides broad temporal coverage at low cost, but can miss brief relevant events when a fixed frame budget is spread across a longer video~\citep{yu2023sevila,tang2025aks}. Query-aware selection concentrates a limited budget on semantically relevant frames~\citep{wang2025videotree,tang2025aks}, but it does not by itself solve the representation problem. Selected frames may be temporally fragmented, and encoding all of them in one context couples the frame budget to the backbone's memory cost. Moreover, query-aware selection changes the retrieval setting: because the selected frames depend on the query, the resulting video representation cannot generally be reused as a single query-independent corpus embedding. A practical system must therefore distinguish query-independent indexing from query-conditioned candidate scoring, as in embedding-and-reranking pipelines~\citep{li2026qwen3vlembedding}.

A second issue concerns how selected evidence is aggregated. Video-retrieval encoders have explored joint spatiotemporal encoding and pooling of frame-level representations~\citep{bain2021frozen,xu2021videoclip,luo2022clip4clip}, while memory-augmented video models explicitly retain information across observations~\citep{song2024moviechat,he2024malmm}. These developments motivate studying how a compact streaming state can be trained to produce useful retrieval embeddings throughout a video. This issue is particularly important when relevant observations are distributed across time or when the number of selected frames still exceeds the context that can be processed economically in one forward pass. We therefore study long-video evidence accumulation through a bounded latent state, building on latent bottlenecks and recurrent memory architectures~\citep{jaegle2021perceiver,bulatov2022rmt,hutchins2022blockrecurrent}.

\begin{figure}[t]
\centering
\includegraphics[width=0.8\linewidth]{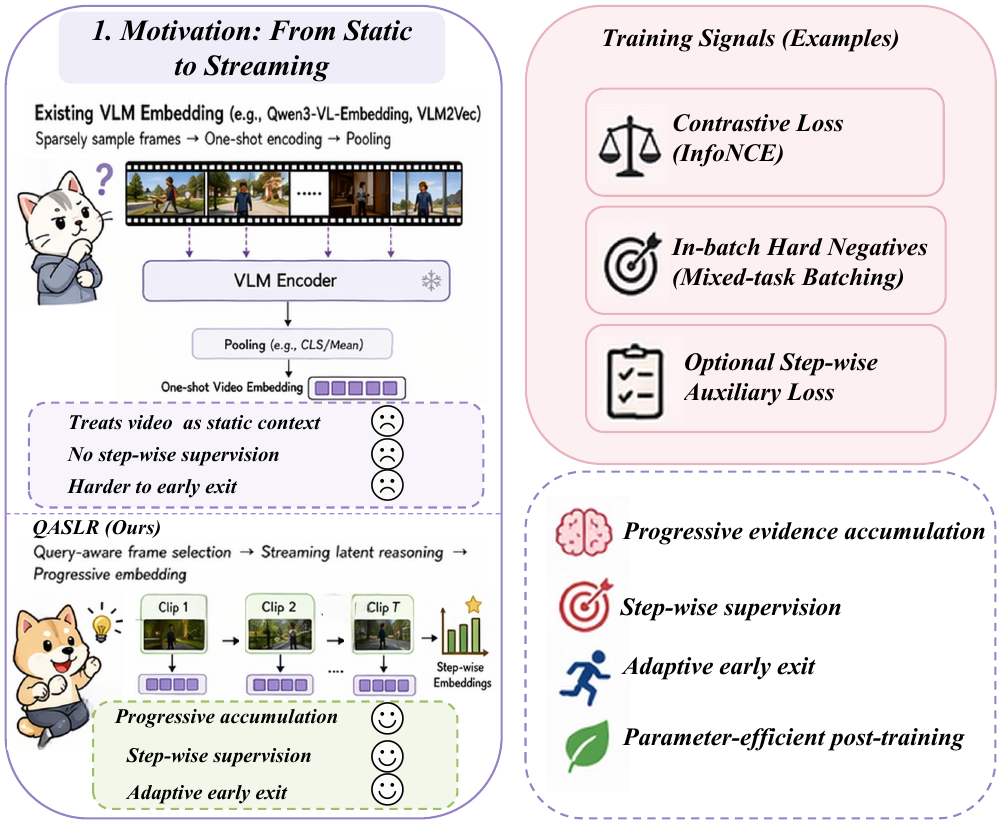}
\caption{\textbf{From one-shot video encoding to streaming evidence accumulation.}
The upper-left pathway illustrates a one-shot baseline that pools representations from sparsely sampled frames into a single video embedding.
The lower-left pathway shows QASLR processing selected frames as ordered clips and producing embeddings throughout the sequence.
The right-hand panels summarize example training signals and the intended capabilities of persistent memory, intermediate supervision, optional early exit, and parameter-efficient post-training.}
\label{fig:rzen_gain_bar_overview}
\end{figure}

We propose \textbf{Query-Aware Streaming Latent Reasoning} (QASLR), a post-training framework for long-video embeddings. As shown in Figure~\ref{fig:rzen_gain_bar_overview}, QASLR processes selected frames as chronologically ordered clips, using learnable think tokens to accumulate clip-level evidence in a persistent latent memory. An embed token reads out a fixed-dimensional representation after each step. Query-independent frame selection enables reusable video indexing, whereas query-aware selection produces query-conditioned representations for a bounded candidate set, such as candidates returned by a first-stage retriever.

Training combines a final contrastive loss with step-wise supervision~\citep{lee2015deeplysupervised} and self-distillation toward the final embedding~\citep{zhang2019beyourownteacher}, aligning both intermediate and complete representations with their targets. Following MMEB and MMEB-V2~\citep{jiang2025vlm2vec,meng2025vlm2vecv2}, a unified query--positive interface supports retrieval, moment retrieval, classification, and multiple-choice video question answering (QA). Interleaved sub-batching groups compatible input formats during mixed-task training and is evaluated separately from streaming aggregation.

The full QASLR training recipe improves both 2B and 8B Qwen3-VL-Embedding backbones~\citep{li2026qwen3vlembedding}, increasing HourVideo retrieval Hit@1 from 54.2 to 70.7 and from 57.4 to 72.8, respectively, relative to the original checkpoints. Gains extend to the evaluated moment-retrieval and video-QA tasks, while experiments on RzenEmbed~\citep{jian2025rzenembed} demonstrate applicability to a second Qwen-family embedding backbone.

Our contributions are threefold:
\begin{itemize}
    \item We introduce QASLR, a streaming framework that decouples temporal coverage from representation size for long-video embedding. A persistent latent memory incrementally integrates temporally ordered clips and emits a fixed-dimensional embedding without placing all selected frames in one backbone context.
    \item We design a streaming post-training objective that combines final contrastive learning, step-wise contrastive supervision, and final-embedding self-distillation. This objective directly supervises intermediate representations through the same query--positive interface used across retrieval, classification, and video QA.
    \item We evaluate QASLR on nine long-video tasks using 2B and 8B Qwen3-VL-Embedding models and demonstrate within-family transfer to RzenEmbed 7B. The full QASLR recipe improves every reported task over the original checkpoints.
\end{itemize}

\section{Related Work}

\subsection{Multimodal Embeddings and Long-Video Processing}

Contrastive learning supports shared image--text and multimodal representations~\citep{radford2021clip,jia2021align,zhai2023siglip,girdhar2023imagebind}, with temporal encoders and aggregation extending this approach to video~\citep{bain2021frozen,xu2021videoclip,luo2022clip4clip}. Instruction-guided retrieval and VLM-based embeddings unify diverse retrieval tasks~\citep{wei2024uniir,jiang2024e5v,jiang2025vlm2vec,zhang2025gme}, with recent models supporting video and document inputs~\citep{meng2025vlm2vecv2,jian2025rzenembed,li2026qwen3vlembedding}. Long-video benchmarks highlight challenges in temporal coverage and reasoning~\citep{fu2025videomme,zhou2025mlvu,wang2025lvbench}. Existing approaches extend context~\citep{zhang2025longva}, compress visual tokens~\citep{shen2025longvu}, aggregate clip descriptions~\citep{zhang2024llovi}, select query-relevant frames~\citep{yu2023sevila,wang2025videotree,tang2025aks}, or allocate frame-wise resolution~\citep{jin2026allocembed}. Temporally grounded datasets provide localized language--video supervision~\citep{krishna2017densecaptioning,gao2017tall,lei2021qvhighlights}. Complementing input selection and compression, QASLR accumulates selected clip evidence into a fixed-dimensional embedding.

\subsection{Streaming Latent Reasoning and Adaptive Computation}

Latent-query architectures compress inputs through cross-attention~\citep{jaegle2021perceiver,jaegle2022perceiverio,alayrac2022flamingo,li2023blip2}, while recurrence and memory tokens carry information across sequence blocks~\citep{dai2019transformerxl,bulatov2022rmt,hutchins2022blockrecurrent}. Video models similarly retain historical information~\citep{song2024moviechat,he2024malmm} and support streaming interaction~\citep{chen2024videollmonline,zhang2024flashvstream}. Related approaches introduce computation before prediction through pause tokens~\citep{goyal2024pause}, continuous hidden states~\citep{hao2025coconut}, or explicit reasoning traces for embeddings~\citep{cui2026thinkthenembed}. Building on these ideas, QASLR updates persistent latent memory across clips and contrastively supervises intermediate embedding readouts. Early-exit methods adapt computation using intermediate predictions~\citep{xin2020deebert,elbayad2020depthadaptive,schuster2022calm}. QASLR's optional exit instead controls the number of observed clips.

\begin{figure}[t]
\centering
\includegraphics[width=\linewidth]{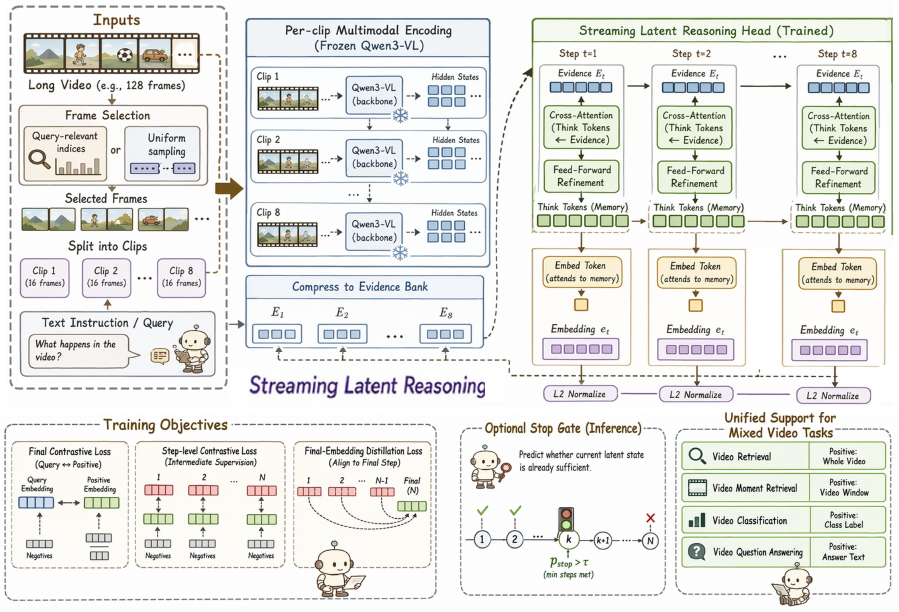}
\caption{\textbf{QASLR architecture, training objectives, and optional early-exit inference.}
Selected frames are ordered and partitioned into clips. The illustrated 128-frame configuration contains eight clips of 16 frames each.
Per-clip backbone features are compressed into evidence banks that update persistent think-token memory through cross-attention and feed-forward refinement.
An embed token reads the memory to produce an L2-normalized embedding after each update.
Training combines final contrastive learning, step-wise supervision, and final-embedding self-distillation across the supported video tasks.
The optional stop gate terminates processing when its stopping criterion is met.}
\label{fig:rzen_gain_bar_detail}
\end{figure}
\section{Method}
\label{sec:method}

\subsection{Overview and Retrieval Settings}

Figure~\ref{fig:rzen_gain_bar_detail} summarizes QASLR. A fixed set of think tokens accumulates clip evidence, and an embed token reads the latent state after each update. Streaming refers to aggregation over preselected clips, not necessarily online frame selection from an unseen video.

Following unified multimodal embedding frameworks~\citep{wei2024uniir,jiang2025vlm2vec,meng2025vlm2vecv2}, we formulate all tasks as query--positive pairs \((A_i,B_i)\). The query \(A_i\) may contain text, an image--text input, or a video--question input, while \(B_i\) may be a video, temporal window, class label, or answer. Non-video and short-video inputs retain the original embedding pathway. Long videos use the proposed streaming pathway.

QASLR supports two retrieval settings, distinguishing reusable dense representations~\citep{karpukhin2020dpr} from pair-conditioned scoring~\citep{li2026qwen3vlembedding}. In the \emph{indexable setting}, frame selection is query-independent, and each candidate video produces one reusable embedding \(\mathbf{v}_j\). In the \emph{query-conditioned setting}, frame selection and video encoding depend on query \(A_i\), producing \(\mathbf{v}_{ij}\). The similarity score is
\begin{equation}
s_{ij}
=
\begin{cases}
\mathbf{q}_i^\top\mathbf{v}_j/\tau,
& \text{indexable},\\[2pt]
\mathbf{q}_i^\top\mathbf{v}_{ij}/\tau,
& \text{query-conditioned},
\end{cases}
\end{equation}
where \(\mathbf{q}_i\) is the normalized query embedding and \(\tau\) is the contrastive temperature. Query-conditioned scoring is restricted to a bounded candidate set, such as the output of a first-stage retriever. Every candidate for a given query is selected and encoded using the same conditioning procedure.

\subsection{Streaming Video Encoding}

Query-guided frame selection has been explored for video QA~\citep{yu2023sevila,tang2025aks}. QASLR treats selection as an input stage: for a long video \(V\) and optional conditioning input \(c\), the selector returns an ordered sequence of at most \(K_{\max}\) indices:
\begin{equation}
\mathcal{I}(V,c)
=
(k_1,\ldots,k_{K_V}),
\qquad
k_1<\cdots<k_{K_V},
\qquad
K_V\leq K_{\max}.
\end{equation}
The selected frames are restored to chronological order and divided into clips:
\begin{equation}
X(V,c)
=
[C_1,\ldots,C_T],
\qquad
T
=
\left\lceil\frac{K_V}{L}\right\rceil,
\end{equation}
where \(L\) is the number of frames per clip. We use \(K_{\max}=128\), with \(L=16\) at full budget and fewer frames per clip for shorter inputs. Padding in the final clip is excluded by the attention mask. Ground-truth labels and temporal boundaries are not provided to the frame selector during evaluation.

At step \(t\), the vision-language backbone \(F_\theta\), instantiated from Qwen3-VL-Embedding~\citep{li2026qwen3vlembedding} built on Qwen3-VL~\citep{bai2025qwen3vl}, encodes the current clip together with the task instruction and optional conditioning input:
\begin{equation}
U_t
=
F_\theta(C_t,c)
\in
\mathbb{R}^{B\times S_t\times D},
\end{equation}
where \(B\), \(S_t\), and \(D\) denote the batch size, valid sequence length, and backbone hidden dimension.

To bound the input size of the streaming head, the valid hidden states are compressed into at most \(M\) evidence tokens:
\begin{equation}
E_t
=
\operatorname{Compress}(U_t)
\in
\mathbb{R}^{B\times M\times D}.
\end{equation}
The compression operator uses cosine similarity to compress valid hidden states after masking. We use \(M=256\) and \(N=32\) think tokens by default. Evidence banks with fewer than \(M\) states are padded and masked in cross-attention. Clips may share a backbone forward pass, but their evidence banks and latent states remain independent.

\subsection{Streaming Latent Memory and Embedding Readout}

Motivated by compact latent bottlenecks~\citep{jaegle2021perceiver} and recurrent memory tokens~\citep{bulatov2022rmt}, QASLR maintains \(N\) learnable think tokens as a persistent latent memory:
\begin{equation}
H_0
\in
\mathbb{R}^{B\times N\times D}.
\end{equation}
The memory is initialized once for each video and is not reset between clips. At streaming step \(t\), the previous memory attends to the current clip evidence using multi-head attention~\citep{vaswani2017attention} and layer normalization~\citep{ba2016layernorm}:
\begin{equation}
\widetilde{H}_t
=
\operatorname{LN}
\left[
H_{t-1}
+
\operatorname{MHA}
\left(
\operatorname{LN}(H_{t-1}),
\operatorname{LN}(E_t),
\operatorname{LN}(E_t)
\right)
\right],
\end{equation}
\begin{equation}
H_t
=
\operatorname{LN}
\left[
\widetilde{H}_t
+
\operatorname{FFN}(\widetilde{H}_t)
\right].
\end{equation}
The recurrent update allows the model to integrate new visual evidence while retaining information from earlier clips. Because the number of think tokens remains fixed, the size of the persistent state does not grow with video duration.

After each update, a learnable embed token \(e\) reads the current memory through query-based attention pooling, as in Perceiver IO~\citep{jaegle2022perceiverio}:
\begin{equation}
r_t
=
\operatorname{MHA}(e,H_t,H_t),
\qquad
\mathbf{z}_t
=
\frac{Wr_t}{\|Wr_t\|_2},
\end{equation}
where \(W\) projects the readout into the shared \(d\)-dimensional embedding space. The streaming pathway returns
\begin{equation}
Z(V,c)
=
(\mathbf{z}_1,\ldots,\mathbf{z}_T),
\end{equation}
and uses \(\mathbf{z}_T\) as the final video representation. Intermediate embeddings expose the representation after each observed clip and are used by the auxiliary training objectives below.

\subsection{Training Objective}

For a batch of \(B\) query--positive pairs, let \(\mathbf{a}_i\) denote the final query-side embedding and \(\mathbf{b}_{ij}\) the representation of candidate \(B_j\) when scored against \(A_i\). In the indexable setting, \(\mathbf{b}_{ij}=\mathbf{b}_j\). In the query-conditioned setting, each candidate is encoded under \(A_i\). The final loss uses an InfoNCE-style contrastive objective~\citep{oord2018cpc,radford2021clip}:
\begin{equation}
\mathcal{L}_{\mathrm{final}}
=
\operatorname{CE}(\widetilde{S},y),
\qquad
s_{ij}
=
\frac{\mathbf{a}_i^\top\mathbf{b}_{ij}}{\tau},
\qquad
y_i=i.
\end{equation}
Query-independent embeddings can be gathered across workers to enlarge the negative pool~\citep{chen2020simclr}. Query-conditioned scores instead require each candidate to be encoded under the current query. Embeddings conditioned on another query are not interchangeable. Valid co-positives from the same video are masked from the denominator, while the paired target remains the sole positive~\citep{khosla2020supcon,jian2025rzenembed}.

Here, \(\widetilde{S}\) equals \(S\) with excluded co-positive logits set to \(-\infty\). Following deep supervision~\citep{lee2015deeplysupervised}, we also train intermediate embeddings. Let \(\mathcal{B}_t\) contain pairs with valid step-\(t\) readouts on all streaming sides. Non-streaming embeddings are reused. The step-wise loss is
\begin{equation}
\mathcal{L}_{\mathrm{step}}
=
\frac{
\sum_t |\mathcal{B}_t|
\operatorname{CE}(\widetilde{S}^{(t)},y^{(t)})
}{
\sum_t |\mathcal{B}_t|
},
\qquad
s_{ij}^{(t)}
=
\frac{
\left(\mathbf{a}_i^{(t)}\right)^\top
\mathbf{b}_{ij}^{(t)}
}{\tau}.
\end{equation}
At each step, invalid pairs and padded readouts are excluded, targets are reindexed within \(\mathcal{B}_t\), and the same co-positive mask is applied. Text-to-video retrieval reuses text embeddings. Video QA reuses answer embeddings.

To regularize the streaming trajectory, we adapt knowledge distillation~\citep{hinton2015distilling} and self-distillation~\citep{zhang2019beyourownteacher} by using the final embedding as a stop-gradient teacher:
\begin{equation}
\mathcal{L}_{\mathrm{distill}}
=
\frac{1}{Z}
\sum_{i\in\mathcal{S}}
\sum_{t=1}^{T_i-1}
\left[
1-
\left(\mathbf{z}_i^{(t)}\right)^\top
\operatorname{sg}
\left(
\mathbf{z}_i^{(T_i)}
\right)
\right],
\end{equation}
where \(\mathcal{S}\) contains the streamed inputs, \(T_i\) is the number of valid steps, and \(Z=\sum_{i\in\mathcal{S}}(T_i-1)\). When \(Z=0\), we define \(\mathcal{L}_{\mathrm{distill}}=0\). Unlike the moving-average target network in BYOL~\citep{grill2020byol}, the teacher here is the detached final-step embedding from the same forward trajectory.

The complete objective is
\begin{equation}
\mathcal{L}
=
\mathcal{L}_{\mathrm{final}}
+
\lambda_{\mathrm{step}}\mathcal{L}_{\mathrm{step}}
+
\lambda_{\mathrm{distill}}\mathcal{L}_{\mathrm{distill}}.
\end{equation}

Mixed-task training combines retrieval, moment retrieval, classification, and video QA through the same query--positive interface. Interleaved sub-batching groups examples with compatible task and input formats. Their losses are weighted by the corresponding sub-batch sizes, and gradients are accumulated before one optimizer update.

Only the streaming head is trainable in the QASLR experiments. The backbone is frozen. The appendix describes data processing, computational complexity, and the optional early-exit mechanism.

\section{Experiments}

\subsection{Experimental Setup}

\paragraph{Tasks and metrics.}
We evaluated nine long-video embedding tasks spanning classification, video retrieval, moment retrieval, and multiple-choice QA. These comprised HourVideo~\citep{chandrasegaran2024hourvideo} classification, retrieval, and moment retrieval, LVBench~\citep{wang2025lvbench} retrieval and QA, LongVideoBench~\citep{wu2024longvideobench} moment retrieval, MovieChat QA~\citep{song2024moviechat}, EgoSchema~\citep{mangalam2023egoschema}, and TimeScope~\citep{zohar2025timescope}. We report Hit@1/Hit@5 for classification and retrieval, except LongVideoBench moment retrieval (Hit@1 only), and accuracy for QA. All reported gains are in percentage points.

\paragraph{Models and training.}
We compared original Qwen3-VL-Embedding~\citep{li2026qwen3vlembedding} checkpoints, single-round post-training, and QASLR at 2B and 8B scales. Both post-training methods used the same training data. QASLR selected at most 128 frames, using 16-frame clips at full budget and shorter clips for fewer frames. Training combined final contrastive learning, step-wise contrastive supervision, and final-embedding self-distillation. Only the streaming head was trainable in QASLR. The backbone remained frozen.

\paragraph{Mixed-task data processing.}
All examples used a common query--positive format. Retrieval paired a query with a video or candidate moment, whereas QA paired a video--question input with its answer text. Selected frames were stored as indices and decoded from source videos on demand. Interleaved sub-batching grouped compatible task and input formats, with its effects examined in Section~\ref{sec:sub_batching}.

\paragraph{Comparison protocol.}
The original checkpoints establish performance before long-video post-training. Single-round post-training uses the mixed-task data without recurrent latent aggregation. Table~\ref{tab:main_qwen3_results} evaluates the complete QASLR recipe rather than isolated architectural changes. Table~\ref{tab:sub_batching_ablation} additionally compares both post-training methods with sub-batching, while Appendix~\ref{app:factorial_ablation} clarifies their interpretation. Training and evaluation details are provided in Appendices~\ref{app:implementation}--\ref{app:retrieval_protocol}.

\begin{table*}[t]
\centering
\caption{\textbf{Long-video embedding performance on Qwen3-VL-Embedding backbones.}
The original checkpoints, single-round post-training (PT), and full QASLR recipe are compared at the 2B and 8B scales.
CLS, Ret., and MRet.\ denote classification, video retrieval, and video moment retrieval, respectively.
Retrieval-style entries report Hit@1/Hit@5, except LongVideoBench MRet., which reports Hit@1 only.
Question-answering (QA) entries report accuracy. Avg.\ is the unweighted mean across the four QA tasks.
All scores are percentages, with higher values indicating better performance.
Bold numeric entries identify the best result for each metric within each backbone scale.}
\label{tab:main_qwen3_results}
\vspace{2pt}
\scriptsize
\setlength{\tabcolsep}{4.2pt}

\begin{tabular}{llccccc}
\toprule
\multicolumn{7}{c}{\textbf{Retrieval-style tasks}} \\
\midrule
Backbone & Method
& HourVideo CLS
& HourVideo Ret.
& HourVideo MRet.
& LVBench Ret.
& LongVideoBench MRet. \\
\midrule
\multirow{3}{*}{Qwen3-VL 2B}
& Original
& 51.1 / 85.1
& 54.2 / 77.1
& 24.3 / 62.2
& 65.4 / 98.1
& 34.2 \\
& Single-round PT
& 52.4 / 86.3
& 55.0 / 76.7
& 25.2 / 63.0
& 66.6 / 98.0
& 35.1 \\
& \textbf{QASLR}
& \textbf{54.2 / 88.1}
& \textbf{70.7 / 92.5}
& \textbf{30.1 / 65.8}
& \textbf{75.7 / 98.3}
& \textbf{37.0} \\
\midrule
\multirow{3}{*}{Qwen3-VL 8B}
& Original
& 52.2 / 83.9
& 57.4 / 87.5
& 24.9 / 63.1
& 68.1 / \textbf{98.4}
& 33.2 \\
& Single-round PT
& 53.0 / 84.3
& 57.8 / 88.4
& 25.6 / 63.8
& 69.2 / 98.1
& 34.5 \\
& \textbf{QASLR}
& \textbf{57.3 / 89.6}
& \textbf{72.8 / 95.8}
& \textbf{27.7 / 66.6}
& \textbf{77.3} / 98.2
& \textbf{38.8} \\
\midrule
\multicolumn{7}{c}{\textbf{QA-style tasks}} \\
\midrule
Backbone & Method
& MovieChat QA
& LVBench QA
& EgoSchema
& TimeScope
& Avg. \\
\midrule
\multirow{3}{*}{Qwen3-VL 2B}
& Original
& 54.1
& 41.1
& 59.8
& 38.0
& 48.3 \\
& Single-round PT
& 54.5
& 41.4
& 60.4
& 39.2
& 48.9 \\
& \textbf{QASLR}
& \textbf{55.2}
& \textbf{42.3}
& \textbf{63.5}
& \textbf{41.3}
& \textbf{50.6} \\
\midrule
\multirow{3}{*}{Qwen3-VL 8B}
& Original
& 55.7
& 38.9
& 70.2
& 80.0
& 61.2 \\
& Single-round PT
& 55.2
& 39.6
& 71.7
& 81.3
& 62.0 \\
& \textbf{QASLR}
& \textbf{60.3}
& \textbf{44.7}
& \textbf{74.5}
& \textbf{85.6}
& \textbf{66.3} \\
\bottomrule
\end{tabular}

\vspace{2pt}
\end{table*}

\begin{table}[t]
\centering
\caption{\textbf{Effects of streaming aggregation and interleaved sub-batching.}
All configurations use query-aware selection. PT and QASLR are evaluated with or without sub-batching.
Ret.\ and MRet.\ denote video retrieval and video moment retrieval.
Retrieval entries report Hit@1/Hit@5, except LongVideoBench MRet., which reports Hit@1 only.
Avg.\ H@1 averages Hit@1 across the three retrieval tasks. QA entries report accuracy.
All scores are percentages, and bold numeric entries mark the best result within each backbone and metric.
QASLR with sub-batching achieves the highest retrieval Hit@1, whereas the strongest QA configuration varies across tasks.}
\label{tab:sub_batching_ablation}
\vspace{2pt}
\scriptsize
\setlength{\tabcolsep}{3.0pt}

\resizebox{\columnwidth}{!}{%
\begin{tabular}{llcccc}
\toprule
\multicolumn{6}{c}{\textbf{Retrieval-style tasks}} \\
\midrule
Backbone
& Method
& HourVideo Ret.
& LVBench Ret.
& LongVideoBench MRet.
& Avg. H@1 \\
\midrule
\multirow{5}{*}{Qwen3-VL 2B}
& Original
& 54.2 / 77.1
& 65.4 / 98.1
& 34.2
& 51.3 \\
& Single-round PT
& 55.0 / 76.7
& 66.6 / 98.0
& 35.1
& 52.2 \\
& Single-round PT w/ sub-batching
& 66.1 / 84.2
& 67.5 / 98.2
& 36.7
& 56.8 \\
& QASLR w/o sub-batching
& 55.4 / 78.8
& 73.3 / 98.0
& 34.8
& 54.5 \\
& \textbf{QASLR w/ sub-batching}
& \textbf{70.7 / 92.5}
& \textbf{75.7 / 98.3}
& \textbf{37.0}
& \textbf{61.1} \\
\midrule
\multirow{5}{*}{Qwen3-VL 8B}
& Original
& 57.4 / 87.5
& 68.1 / \textbf{98.4}
& 33.2
& 52.9 \\
& Single-round PT
& 57.8 / 88.4
& 69.2 / 98.1
& 34.5
& 53.8 \\
& Single-round PT w/ sub-batching
& 67.3 / 93.5
& 75.8 / 98.0
& 36.7 
& 59.9\\
& QASLR w/o sub-batching
& 59.2 / 89.9
& 72.4 / 98.3
& 35.6
& 55.7 \\
& \textbf{QASLR w/ sub-batching}
& \textbf{72.8 / 95.8}
& \textbf{77.3} / 98.2
& \textbf{38.8}
& \textbf{63.0} \\
\midrule
\multicolumn{6}{c}{\textbf{QA-style tasks}} \\
\midrule
Backbone
& Method
& MovieChat
& LVBench
& EgoSchema
& TimeScope \\
\midrule
\multirow{5}{*}{Qwen3-VL 2B}
& Original
& 54.1
& 41.1
& 59.8
& 38.0 \\
& Single-round PT
& 54.5
& 41.4
& 60.4
& 39.2 \\
& Single-round PT w/ sub-batching
& \textbf{58.8}
& \textbf{47.2}
& 62.9
& \textbf{42.4} \\
& QASLR w/o sub-batching
& 47.0
& 36.8
& 60.7
& 38.8 \\
& \textbf{QASLR w/ sub-batching}
& 55.2
& 42.3
& \textbf{63.5}
& 41.3 \\
\midrule
\multirow{5}{*}{Qwen3-VL 8B}
& Original
& 55.7
& 38.9
& 70.2
& 80.0 \\
& Single-round PT
& 55.2
& 39.6
& 71.7
& 81.3 \\
& Single-round PT w/ sub-batching
& \textbf{61.5}
& 43.3
& 73.8
& 84.1 \\
& QASLR w/o sub-batching
& 53.2
& 41.7
& 72.2
& 80.5 \\
& \textbf{QASLR w/ sub-batching}
& 60.3
& \textbf{44.7}
& \textbf{74.5}
& \textbf{85.6} \\
\bottomrule
\end{tabular}%
}
\vspace{2pt}
\end{table}

\begin{figure}[t]
\centering
\includegraphics[width=\linewidth]{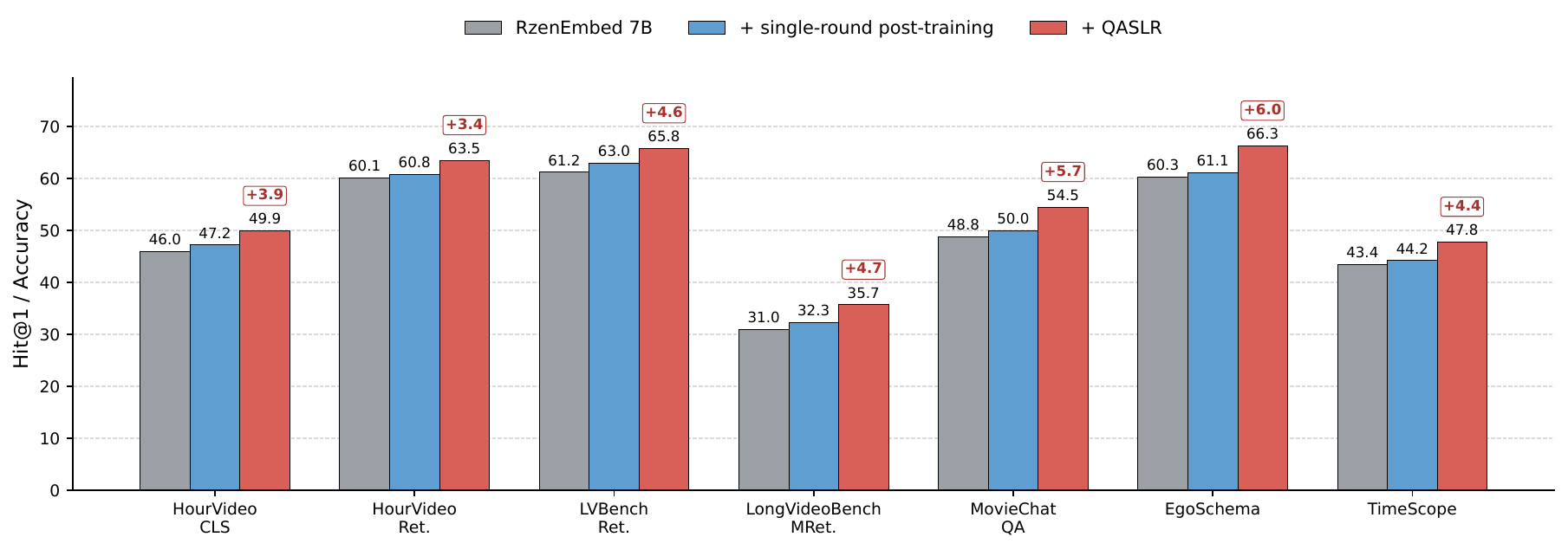}
\caption{\textbf{Within-family transfer of QASLR to RzenEmbed 7B.}
Grouped bars compare the original checkpoint, single-round post-training, and QASLR across seven tasks using the same RzenEmbed backbone.
Classification (CLS), video retrieval (Ret.), and moment retrieval (MRet.) report Hit@1. Question-answering tasks report accuracy.
All scores are percentages, and annotations above the QASLR bars give absolute improvements over the original checkpoint in percentage points.
QASLR exceeds both baselines on every displayed task, with gains of 3.4--6.0 points over the original checkpoint.
}
\label{fig:rzen_gain_bar}
\end{figure}

\subsection{Main Results on Qwen3-VL-Embedding}

In Table~\ref{tab:main_qwen3_results}, QASLR achieved the highest Hit@1 on all five retrieval-style tasks at both scales. Mean Hit@1 increased from 45.8 to 53.5 at 2B and from 47.2 to 54.8 at 8B, compared with single-round averages of 46.9 and 48.0. The similar gains over the original checkpoints show that improvement was not limited to one model scale.

\paragraph{Video retrieval and classification.}
The largest gains occurred on full-video retrieval. QASLR achieved 70.7/92.5 HourVideo Hit@1/Hit@5 at 2B and 72.8/95.8 at 8B, exceeding the original checkpoints by 16.5 and 15.4 Hit@1 points, respectively. Relative to single-round post-training, the corresponding Hit@1 gains were 15.7 and 15.0 points. HourVideo Hit@5 also improved by 15.4 and 8.3 points over the original checkpoints, showing gains in both top-one and top-five retrieval. On LVBench retrieval, Hit@1 improved by 10.3 and 9.2 points over the original checkpoints, while Hit@5 remained near 98. Thus, the LVBench gains primarily reflect better top-ranked ordering rather than expanded top-five coverage. HourVideo classification improved more moderately, from 51.1 to 54.2 at 2B and from 52.2 to 57.3 at 8B.

\paragraph{Moment retrieval.}
HourVideo moment-retrieval Hit@1 increased from 24.3 to 30.1 at 2B and from 24.9 to 27.7 at 8B, with Hit@5 also improving at both scales. LongVideoBench Hit@1 rose from 34.2 to 37.0 and from 33.2 to 38.8, respectively. The larger HourVideo gain at 2B and larger LongVideoBench gain at 8B indicate task-dependent effects of model scale.

\paragraph{Video question answering.}
QASLR improved all four QA tasks over both baselines in Table~\ref{tab:main_qwen3_results}. Mean accuracy increased from 48.3 to 50.6 at 2B and from 61.2 to 66.3 at 8B relative to the original checkpoints. The 8B gains were 4.6 points on MovieChat, 5.8 on LVBench, 4.3 on EgoSchema, and 5.6 on TimeScope. Improvements at 2B ranged from 1.1 to 3.7 points. Single-round post-training achieved mean QA accuracies of 48.9 and 62.0, so the additional gains with QASLR were larger at 8B. These results concern the complete recipe. Comparison with sub-batched single-round training reveals a less uniform QA advantage, as discussed below.

\subsection{Generalization to Another VLM Embedding Backbone}

Figure~\ref{fig:rzen_gain_bar} compares the original RzenEmbed 7B checkpoint, single-round post-training, and QASLR. This Qwen2-VL-based backbone provides a test beyond Qwen3-VL-Embedding using the same training formulation. QASLR exceeded both baselines on all seven displayed tasks, with gains of 3.4--6.0 points over the original checkpoint. Improvements span classification, retrieval, moment retrieval, and QA, rather than a single task family. Relative to single-round post-training, QASLR raised EgoSchema accuracy from 61.1 to 66.3 and LongVideoBench moment-retrieval Hit@1 from 32.3 to 35.7. Because both backbones belong to the Qwen family, this experiment supports within-family transfer, not generalization across unrelated architectures.

\subsection{Effect of Interleaved Sub-Batching}
\label{sec:sub_batching}

Table~\ref{tab:sub_batching_ablation} shows that the sub-batching configuration, including a larger effective batch, improved QASLR's mean retrieval Hit@1 from 54.5 to 61.1 at 2B and from 55.7 to 63.0 at 8B. HourVideo retrieval gains were 15.3 and 13.6 points, respectively. LVBench retrieval improved by 2.4 and 4.9 Hit@1 points, while LongVideoBench moment retrieval improved by 2.2 and 3.2 points. At 2B, MovieChat and LVBench QA fell below the original checkpoint without sub-batching, reaching 47.0 and 36.8. Sub-batching recovered them to 55.2 and 42.3. At 8B, QA gains ranged from 2.3 points on EgoSchema to 7.1 on MovieChat. Sub-batching therefore benefited both retrieval and QA within the QASLR configuration.

However, sub-batching also improved single-round post-training: mean retrieval Hit@1 increased from 52.2 to 56.8 at 2B and from 53.8 to 59.9 at 8B. With sub-batching, QASLR retained higher Hit@1 on all three retrieval tasks at both scales. Mean scores were 61.1 versus 56.8 at 2B and 63.0 versus 59.9 at 8B. At the task level, HourVideo Hit@1 gains over this baseline were 4.6 points at 2B and 5.5 at 8B. The corresponding margins were 8.2 and 1.5 points on LVBench, and 0.3 and 2.1 on LongVideoBench, showing task- and scale-dependent differences. QA results were mixed: single-round post-training scored higher on three of four tasks at 2B and on MovieChat at 8B. These comparisons show substantial gains from the batching configuration, while QASLR's advantage over the sub-batched baseline is consistent for retrieval but task-dependent for QA.

\section{Conclusion}

We introduced QASLR, a post-training framework that represents long videos through a fixed-size latent memory updated over temporally ordered clips. The full training recipe improved retrieval, moment retrieval, classification, and multiple-choice QA relative to the original checkpoints at both the 2B and 8B scales. The streaming head also transferred to a second Qwen-family embedding model. QASLR stores frame indices rather than duplicated images and preserves the downstream embedding interface. Its query-aware path produces query-conditioned representations and is therefore intended for bounded candidate sets or second-stage reranking. Query-independent selection remains necessary when each video must be indexed once and reused across queries. Sub-batching benefits both single-round and streaming models, although improvements in QA remain task-dependent. Future work could assess generalization beyond Qwen-family backbones and evaluate whether adaptive stopping can reduce inference cost while preserving task performance.

\subsection*{AI use statement}
Generative AI tools were used solely to improve the clarity and readability
of the manuscript. They were not used to formulate scientific claims,
design experiments, generate results, or interpret findings. All
AI-assisted edits were reviewed by the authors, who take full
responsibility for the manuscript.

\subsection*{Ethics Statement}
This work does not involve human participants, personally identifiable
information, or private data. We recognize that the proposed method may
inherit biases and limitations from its training data and underlying
models. Potential misuse and unintended deployment risks should therefore
be carefully considered in real-world applications. All authors have
reviewed the ethical implications of this work and take responsibility
for its content.

\subsection*{Reproducibility Statement}
We provide detailed descriptions of the proposed method, model
configuration, training procedure, datasets, evaluation protocols, and
hyperparameters in the main paper and appendices. We report the
computational resources and random seeds used in our experiments where
applicable. The implementation details necessary to reproduce the main results are
provided in the paper and appendices.

\bibliography{references}
\bibliographystyle{iclr2027_conference}

\clearpage
\appendix
\section{Appendix}

This appendix describes the training configuration, evaluation data, component comparisons, and theoretical properties of QASLR.

\subsection{Detailed Algorithm}
Table~\ref{alg:qaslr_training} summarizes the QASLR training pathway. Each example is represented through a query--positive interface, and selected frame indices are decoded on demand. Streaming memory is initialized once for each input and retained across its clips.

\begin{table}[h]
\centering
\caption{\textbf{QASLR post-training procedure for unified query--positive pairs.}
Selected frame indices are decoded on demand and restored to chronological order before clip-wise backbone encoding.
A bounded evidence bank updates the persistent think-token memory, which is initialized once per video and read out after every clip.
The optimization combines final contrastive learning with the enabled auxiliary objectives.
Interleaved sub-batching groups compatible task and input formats before parameter updates.
The gate objective applies only when training the optional early-exit module; it is not required for fixed-length streaming inference.}
\label{alg:qaslr_training}
\scriptsize
\begin{tabular}{p{0.96\linewidth}}
\toprule
\textbf{Input:} training pairs \(\{(q_i, p_i)\}_{i=1}^{P}\), selected frame indices, VLM backbone \(F_\theta\), streaming head \(G_\phi\), temperature \(\tau\), loss weights \(\lambda_{\mathrm{step}}, \lambda_{\mathrm{distill}}\). \\
\midrule
1. Sample a mini-batch of query-positive pairs. If interleaved sub-batching is enabled, group examples with similar task and input format into homogeneous sub-batches. \\
2. For each streaming video item, decode selected frame indices from the original video and restore chronological order. \\
3. Split selected frames into clips \([C_1,\ldots,C_T]\). \\
4. For each clip \(C_t\), encode \(U_t=F_\theta(C_t,c)\), where \(c\) is the permitted conditioning input, omitted for query-independent video encoding. \\
5. Compress valid hidden states into a fixed-size evidence bank \(E_t\). \\
6. Initialize think-token memory \(H_0\) and update it sequentially: \(H_t = \mathrm{Update}_{\phi}(H_{t-1}, E_t)\). \\
7. Read out step embeddings \(z_t=\mathrm{Readout}_{\phi}(H_t)\). \\
8. Compute the masked final contrastive loss and enabled auxiliary embedding losses; train the optional gate separately after embedding training. \\
9. Update the streaming-head parameters while keeping the pretrained backbone frozen. \\
\bottomrule
\end{tabular}
\end{table}

\paragraph{State isolation and padding.}
Each video trajectory has an independent memory state. Packing clips into a backbone batch does not permit attention or memory sharing between different inputs. Padded evidence tokens and unavailable readouts are masked. Non-streaming inputs retain their original embedding pathway.

\paragraph{Inference.}
Fixed-length inference returns the final valid readout. Optional early exit returns an intermediate readout when the stopping rule is satisfied. Early-exit use in local-clip QA is discussed separately in Section~\ref{app:early_exit_efficiency}.

\subsection{Implementation Details}
\label{app:implementation}

\paragraph{Backbone input and output.}
The image processor uses a pixel-area range of 4,096--1,310,720, with height and width aligned to multiples of 32. The output embedding dimensions are 2,048 for Qwen3-VL-Embedding 2B and 4,096 for Qwen3-VL-Embedding 8B.

\paragraph{Streaming configuration.}
The main QASLR configuration selects at most 128 frames and uses 32 think tokens and at most 256 evidence tokens per clip. At the full frame budget, 16 frames per clip give eight streaming updates. For inputs with fewer selected frames, the number of frames per clip is reduced. Only valid frames and readouts contribute to attention and auxiliary losses. Compression uses cosine similarity between valid hidden-state representations to construct the evidence bank.

\paragraph{Trainable modules.}
Only the streaming latent reasoning head is updated in QASLR, while pretrained backbone parameters remain frozen. The trainable parameter counts are approximately 200 million for the 2B configuration and 600 million for the 8B configuration. These counts refer to the trainable head rather than the total model size.

\paragraph{Optimization.}
Post-training uses AdamW with a learning rate of $10^{-4}$ and approximately three to five epochs, depending on the configuration. Training runs on NVIDIA H100 GPUs, with a per-GPU batch size of 4 or 8 and a global effective batch size of 256 or 512 across configurations. The step-wise contrastive and self-distillation weights are $\lambda_{\mathrm{step}}=0.1$ and $\lambda_{\mathrm{distill}}=0.01$, respectively. The reported experimental results are averaged over three random seeds.

\paragraph{Interleaved sub-batching.}
Compatible examples are grouped by task and input format. Negative candidates are shared across sub-batches and GPUs, and the effective batch size increases in the sub-batching configuration. The sub-batching configuration therefore changes both the effective optimization batch size and the negative pool. For query-conditioned scoring, shared candidates retain the query-conditioning requirement defined in the main objective.

\paragraph{Frame storage.}
Selected frames are recorded as indices with optional timestamps and decoded from the original videos when needed. This avoids storing duplicated image copies for every query--video annotation. Frame indices and timestamps provide references to the original video without duplicating the selected images.

\subsection{Additional Experimental Details}
\label{app:data_protocol}

\paragraph{Post-training data.}
Post-training primarily uses LLaVA-Video-178K, supplemented with 10,000 ActivityNet videos.

\paragraph{Benchmark annotations and task construction.}
QA annotations are taken from the corresponding original benchmark datasets. The four QA results in the main comparison are MovieChat, LVBench, EgoSchema, and TimeScope. Classification, video retrieval, and moment retrieval use sampled and filtered subsets of the source datasets. Their sizes vary by dataset and task, from tens to thousands of examples. GPT-4o assists annotation by constructing pairs of visual frames from video segments and textual questions. We will publicly release the complete evaluation sets for all reported tasks to support reproducibility and future research on long-video understanding.

\paragraph{Candidate sets.}
These evaluations use bounded candidate sets. Video-retrieval questions generally have 4 or 8 candidates, and moment-retrieval questions generally have 8 or 16 candidate segments, depending on video length. Classification uses 4--8 categories, depending on the dataset. QA questions generally contain four answer options. Retrieval scores measure ranking within these task-specific candidate sets. The candidate-count distribution also affects Hit@5: with four candidates and a valid positive, top-five retrieval necessarily includes that positive.

\paragraph{QA scoring.}
The video--question input is embedded on the query side, and the original answer options are embedded on the candidate side. The selected answer is the option with the highest embedding similarity. No additional QA prompt is introduced beyond this input construction.

\paragraph{Metrics and averaging.}
Retrieval-style tasks report Hit@1/Hit@5 where available, and QA reports accuracy. LongVideoBench moment retrieval reports Hit@1. Results are averaged over three random seeds. Task averages include only the tasks specified in the relevant table and do not weight by the number of examples.

\subsection{Evaluation Modes and Query Conditioning}
\label{app:retrieval_protocol}

\paragraph{Query-aware frame selection.}
We use Q-Frame for training-free frame selection conditioned on text. For a given retrieval query, the same query conditions selection and encoding for both positive and negative candidates. Ground-truth labels and annotated temporal boundaries are not provided to the selector during evaluation. The resulting video representation is query-conditioned and is intended for bounded candidate scoring.

\paragraph{Indexing and two-stage scoring.}
Query-independent selection and encoding produce one reusable representation per video. Query-aware selection instead requires candidate processing after the query is available. Table~\ref{tab:query_conditioning_protocol} distinguishes these deployment modes. In its two-stage comparison, the single-round and QASLR methods use the same first-stage candidate lists, and first-stage misses remain in the denominator of final Hit@1. Table~\ref{tab:sub_batching_ablation} uses query-aware selection throughout, whereas Table~\ref{tab:query_conditioning_protocol} additionally evaluates query-independent indexing.

\begin{table}[t]
\centering
\caption{\textbf{Indexable and query-conditioned retrieval on HourVideo with Qwen3-VL-Embedding 2B.}
Uniform-selection methods produce one query-independent representation per video that can be reused across queries.
SB denotes interleaved sub-batching; the uniform single-round row does not use SB.
Two-stage methods apply query-aware selection and candidate scoring only to the top-\(M\) videos returned by a query-independent first stage.
First-stage R@\(M\) measures positive-video recall before second-stage scoring, and Hit@1 measures final retrieval performance; both are percentages.
\emph{First stage} indicates that only the initial retrieval stage is pre-indexable, and a dash marks an inapplicable shortlist-recall entry.
Latency is measured in seconds per query on NVIDIA H100 hardware with a warm cache.}
\label{tab:query_conditioning_protocol}
\scriptsize
\setlength{\tabcolsep}{3.5pt}
\begin{tabular}{lccccc}
\toprule
Mode & Query-aware & Pre-indexable & First-stage R@$M$ & Hit@1 & Latency (s) \\
\midrule
Uniform, single-round
& No & Yes & -- & 55.0 & 0.18 \\
Uniform, QASLR + SB
& No & Yes & -- & 64.1 & 0.61 \\
Two-stage, single-round + SB
& Yes & First stage & 95.8 & 66.1 & 1.26 \\
Two-stage, QASLR + SB
& Yes & First stage & 95.8 & \textbf{70.7} & 1.72 \\
\bottomrule
\end{tabular}
\end{table}

\paragraph{First-stage recall and final accuracy.}
Let $R$ denote shortlist size, to distinguish it from the evidence-bank size $M$. Let $\mathcal C_R(q)$ be the shortlist and $\mathcal P(q)$ the valid positive set. First-stage recall measures whether $\mathcal C_R(q)\cap\mathcal P(q)$ is nonempty, while final Hit@1 measures whether the top-ranked candidate is positive. A first-stage miss contributes zero to the latter. Consequently, end-to-end Hit@1 cannot exceed first-stage recall. The reported values imply approximate top-one success rates conditional on a first-stage hit of $66.1/95.8=69.0\%$ and $70.7/95.8=73.8\%$. Both conditional rates are calculated from the rounded values in Table~\ref{tab:query_conditioning_protocol}.

\subsection{Attribution of the Main Gain}
\label{app:factorial_ablation}

\paragraph{Training configuration.}
All configurations in Table~\ref{tab:sub_batching_ablation}, including the original checkpoints, use query-aware frame selection. At 2B, single-round HourVideo Hit@1 increases from 55.0 to 66.1 with the sub-batching configuration. This configuration combines task-compatible grouping, cross-sub-batch and cross-GPU negatives, and a larger effective batch size. The comparison measures their combined contribution to post-training.

\paragraph{Streaming contribution.}
Relative to the original checkpoints, the complete QASLR recipe improves HourVideo Hit@1 by 16.5 and 15.4 points at 2B and 8B. Against single-round training with sub-batching, the corresponding gains are 4.6 and 5.5 points. This comparison evaluates the streaming training configuration with sub-batching present in both methods.

\paragraph{Task-dependent QA gains.}
At 2B, QASLR minus sub-batched single-round differences are $-3.6$, $-4.9$, $+0.6$, and $-1.1$ points on MovieChat, LVBench, EgoSchema, and TimeScope. At 8B, the corresponding differences are $-1.2$, $+1.4$, $+0.7$, and $+1.5$ points. Across these comparisons, the retrieval advantage is more consistent than the QA advantage.

\subsection{Does the Latent Memory Use Temporal Evidence?}
\label{app:temporal_controls}

The aggregation controls retain the selected frames, backbone features, training setup, and objectives except for the component being ablated. Mean pooling, a recurrent aggregator, and a Transformer aggregator are compared with QASLR. Clip shuffling, reversal, and memory reset are applied only at test time, without retraining. The last-clip-only control retains the final clip of the selected chronological frame sequence and also does not retrain the model.

\begin{table}[t]
\centering
\caption{\textbf{Aggregation, clip-order, and memory controls with Qwen3-VL-Embedding 2B.}
Variants use the same selected frames and backbone features, except that the last-clip-only control retains only the final clip.
Alternative aggregators test pooling and additional modeling capacity, while QASLR controls perturb clip order or remove cross-clip memory.
\emph{Ordered} indicates chronological clip presentation, not whether an aggregator is sensitive to order; \emph{Persistent memory} indicates retention of information across clips.
Ret.\ Hit@1 is HourVideo retrieval Hit@1, and QA Acc.\ is EgoSchema accuracy.
Scores are percentages, and bold numeric entries mark the highest value in each metric.}
\label{tab:temporal_memory_controls}
\scriptsize
\setlength{\tabcolsep}{3.6pt}
\begin{tabular}{lcccc}
\toprule
Aggregator & Ordered & Persistent memory & Ret. Hit@1 & QA Acc. \\
\midrule
Mean pooling
& Yes & No & 58.6 & 60.2 \\
Recurrent aggregator
& Yes & Yes & 63.5 & 61.8 \\
Transformer aggregator
& Yes & Yes & 66.2 & 62.7 \\
QASLR, shuffled clips
& No & Yes & 64.1 & 61.6 \\
QASLR, reversed clips
& No & Yes & 65.0 & 61.9 \\
QASLR, reset memory
& Yes & No & 60.3 & 60.8 \\
QASLR, last clip only
& Yes & No & 56.9 & 59.7 \\
QASLR
& Yes & Yes & \textbf{70.7} & \textbf{63.5} \\
\bottomrule
\end{tabular}
\end{table}

Relative to full QASLR, HourVideo Hit@1 decreases by 6.6 points with shuffled clips, 5.7 with reversed clips, 10.4 with memory reset, and 13.8 with only the last clip. QASLR exceeds the reported Transformer aggregator by 4.5 points. These controls support the usefulness of cross-clip state and chronological inputs under the evaluated settings. Because the perturbations are applied only at inference, the effects also include sensitivity to a change in the input or state-update distribution.

\subsection{Frame Budget and Streaming Depth}
\label{app:budget_steps}

The full-budget configuration uses $K=128$ frames, $L=16$ frames per clip, and eight memory updates. For inputs with fewer selected frames, QASLR reduces the number of frames per clip. Frame budget and clip size determine different aspects of processing: $K$ controls the observations supplied to the backbone, while $L$ controls per-call context length.

For fixed $K$, changing $L$ also changes the number of updates $T=\lceil K/L\rceil$. At fixed $L$, increasing $K$ changes both temporal coverage and total computation. These dependencies distinguish the storage occupied by the persistent memory from the cost of obtaining and encoding its evidence.

\subsection{Objective Ablations and Intermediate Embedding Quality}
\label{app:objective_ablation}

We isolate the effects of step-wise contrastive supervision and final-embedding self-distillation using a \(2\times2\) objective ablation. The final contrastive loss is active in every configuration, and all other settings are identical across the four loss combinations. For text-to-video retrieval, the text query embedding is shared across steps and contrasted with each intermediate video embedding. For examples with unequal numbers of steps, padded steps are excluded from both the contrastive and distillation losses.

Table~\ref{tab:objective_ablation} reports Hit@1 after 25\%, 50\%, 75\%, and all selected clips. Mean checkpoint Hit@1 is the unweighted average of these four scores, summarizing intermediate and final retrieval quality without treating the checkpoints as an integrated curve.

\begin{table}[t]
\centering
\caption{\textbf{Auxiliary-objective ablation and intermediate retrieval quality on HourVideo.}
The four Qwen3-VL-Embedding 2B configurations independently enable step-wise contrastive supervision and final-embedding self-distillation, with the final contrastive loss active throughout.
Entries report retrieval Hit@1 (\%) after processing 25\%, 50\%, 75\%, or all selected clips; \emph{Final} denotes the complete sequence.
\emph{Mean H@1} is the unweighted mean of the four checkpoint scores.
Bold numeric entries mark the highest value in each column, including intermediate and final performance.}
\label{tab:objective_ablation}
\scriptsize
\setlength{\tabcolsep}{3.3pt}
\begin{tabular}{llccccc}
\toprule
Step loss & Distillation & 25\% & 50\% & 75\% & Final & Mean H@1 \\
\midrule
No
& No
& 43.8 & 54.6 & 62.8 & 67.9 & 57.3 \\
Yes
& No
& 49.7 & 59.8 & 66.5 & 69.8 & 61.5 \\
No
& Yes
& 51.2 & 61.7 & 67.9 & 70.1 & 62.7 \\
Yes
& Yes
& \textbf{55.6} & \textbf{64.8} & \textbf{69.1}
& \textbf{70.7} & \textbf{65.1} \\
\bottomrule
\end{tabular}
\end{table}

\paragraph{Negative construction.}
Consistent with the main objective, each query retains its paired target as the positive. Other valid co-positives from the same source video are excluded from the contrastive denominator, rather than added to a multi-positive target. The mask removes valid co-positives from the denominator while preserving the paired target.

\paragraph{Final and intermediate effects.}
Adding step-wise supervision alone increases final HourVideo Hit@1 from 67.9 to 69.8; distillation alone increases it to 70.1; using both reaches 70.7. The corresponding changes at the 25\% checkpoint are $+5.9$, $+7.4$, and $+11.8$ points. Relative to final-only supervision, the mean over the four checkpoints increases from 57.3 to 65.1 with both objectives. The combined final gain is 2.8 points, with larger improvements at the early checkpoint. Both objectives therefore improve partial-video readouts as well as the final embedding in this comparison.

\paragraph{Partial evidence and distillation.}
An early clip may not yet contain the evidence needed for the paired target. Step-wise supervision therefore encourages a useful partial-video representation, rather than certifying that the answer is already observable. The final teacher also sees evidence unavailable at earlier steps. Alignment to that teacher is a training regularizer, not access to unseen frames at inference. Neither a high checkpoint accuracy nor a low average distillation loss guarantees correctness on every query or every candidate.

\subsection{Frame Selection and Evidence Coverage}
\label{app:selector_robustness}

Q-Frame selects text-relevant frames within the input budget, and QASLR restores their chronological order before aggregation. The selector determines the available observations; the persistent memory combines their evidence across clips. Events omitted by selection remain outside the information available to the streaming head.

\subsection{Early Exit and Computational Cost}
\label{app:early_exit_efficiency}

\paragraph{Optional early exit.}
For local-clip QA, the optional stopping rule returns an intermediate readout at the first eligible step $t\ge t_{\min}$ satisfying $p_t\ge\eta$. If no eligible step meets the threshold, processing continues to the final readout. The fixed-length retrieval results use the complete sequence. The theoretical analysis below characterizes agreement with full-step predictions under its stated calibration assumption.

\paragraph{Warm-cache latency.}
Table~\ref{tab:query_conditioning_protocol} reports per-query latency on NVIDIA H100 hardware with a warm cache. The two-stage single-round and QASLR configurations take 1.26~s and 1.72~s, respectively. QASLR improves HourVideo Hit@1 by 4.6 points in this comparison, with a latency increase of 0.46~s, approximately 36.5\%. The uniform-selection rows compare configurations that additionally differ in sub-batching.

\paragraph{Cost decomposition.}
For a shortlist $\mathcal C_R(q)$ containing at most $R$ candidates, online processing comprises query encoding, search, candidate processing, and ranking:
\begin{align}
C_{\mathrm{online}}(q)
&=C_{\mathrm{query}}(q)+C_{\mathrm{search}}(q)+C_{\mathrm{rank}}(q)\nonumber\\
&\quad+\sum_{j\in\mathcal C_R(q)}
\left[C_{\mathrm{select}}(q,j)+C_{\mathrm{decode}}(q,j)
+C_{\mathrm{encode}}(q,j)+C_{\mathrm{head}}(q,j)\right].
\end{align}
This decomposition defines the full online workload separately from the warm-cache measurements above. Query-independent corpus encoding and index construction contribute offline costs, amortized as $C_{\mathrm{offline}}/Q$ over $Q$ queries. Cached media can reduce repeated decoding, whereas query-conditioned video encoding remains specific to each query--candidate pair.

\subsection{Data Separation and Result Aggregation}
\label{app:reliability}

\paragraph{Source-video separation.}
The post-training and evaluation sets have zero overlap in source videos. This separation concerns the constructed post-training corpus and evaluation sets. QA uses original benchmark annotations, while classification and retrieval use the task-specific subsets described in Section~\ref{app:data_protocol}.

\paragraph{Result aggregation.}
Experimental results are averaged over three random seeds. Differences between configurations are expressed in percentage points, and task-level averages use the metrics and task groups identified in each table. Retrieval metrics are interpreted within the candidate sets defined for each task.

\subsection{Complexity Analysis}

Let \(K\) be the selected-frame count, \(L\) the clip size, \(T=\lceil K/L\rceil\) the number of steps, and \(D\) the backbone hidden dimension. The head uses \(N\) think tokens, at most \(M\) evidence tokens, and an output dimension \(d\). Let \(C_{\mathrm{vlm}}(L)\) denote one clip-encoding cost. For the cross-attention, fixed-width-ratio feed-forward network, and readout defined in Section~\ref{sec:method}, the per-step head cost is
\begin{equation}
C_{\mathrm{head}}
=
O\!\left(NMD+(N+M)D^2+Dd\right).
\end{equation}
The terms account for attention mixing, token projections and feed-forward transformations, and the output projection, respectively. This expression does not assume an additional latent self-attention layer. For full-step inference,
\begin{equation}
C_{\mathrm{full}}
=
C_{\mathrm{fixed}}
+
T\left[C_{\mathrm{vlm}}(L)+C_{\mathrm{head}}\right],
\end{equation}
where \(C_{\mathrm{fixed}}\) includes any up-front selection or preprocessing cost. In query-conditioned retrieval, clip encoding and aggregation are repeated for each query--candidate pair; first-stage retrieval and query encoding are additional costs.

The persistent memory occupies \(O(ND)\) values per example. Excluding model parameters and backbone activations, a conventional single-step attention implementation requires working space of order
\begin{equation}
O\!\left((N+M)D+NM+d\right).
\end{equation}
These are inference-state bounds, not bounds on training peak memory. Backpropagation through the streaming head may retain activations across steps even when the backbone is frozen; checkpointing or recomputation changes that trade-off.

If the model exits after \(\widehat{T}\le T\) steps, then, for fixed per-step cost and negligible gate overhead,
\begin{equation}
\mathbb{E}[C_{\mathrm{exit}}]
=
C_{\mathrm{fixed}}
+
\mathbb{E}[\widehat{T}]
\left[C_{\mathrm{vlm}}(L)+C_{\mathrm{head}}\right].
\end{equation}
When fixed costs are negligible, the remaining cost ratio is approximately \(\mathbb{E}[\widehat{T}]/T\), and the saved fraction is \(1-\mathbb{E}[\widehat{T}]/T\). Up-front frame selection and decoding can reduce end-to-end savings relative to this approximation.

\subsection{Theoretical Analysis}

The following conditional results describe perturbation stability and agreement with full-step retrieval. They do not establish that optimization attains the stated assumptions or that QASLR is optimal.

\paragraph{Assumption 1: regular memory update and readout.}
Let \(H_t=U_\phi(H_{t-1},E_t)\). On the states under consideration, assume
\begin{equation}
\|U_\phi(H,E)-U_\phi(H',E')\|_F
\le
\rho\|H-H'\|_F+\beta\|E-E'\|_F,
\end{equation}
with \(\rho,\beta\ge0\), and assume the normalized readout is \(L_R\)-Lipschitz. For normalization, this requires the pre-normalization readout norm to remain bounded away from zero, in addition to regularity of the preceding operations.

\paragraph{Proposition 1: propagation of evidence perturbations.}
Consider two executions with the same initial memory and \(\|E_t-\widetilde E_t\|_F\le\epsilon\) at every step. Under Assumption 1,
\begin{equation}
\|H_T-\widetilde H_T\|_F
\le
\beta\epsilon\sum_{k=0}^{T-1}\rho^k,
\qquad
\|z_T-\widetilde z_T\|_2
\le
L_R\beta\epsilon\sum_{k=0}^{T-1}\rho^k.
\end{equation}

\paragraph{Proof.}
Writing \(\delta_t=\|H_t-\widetilde H_t\|_F\) gives
\(\delta_t\le\rho\delta_{t-1}+\beta\epsilon\).
Unrolling from \(\delta_0=0\) yields the memory bound; applying readout Lipschitz continuity yields the embedding bound. \(\square\)

If \(\rho<1\), the memory bound is at most \(\beta\epsilon/(1-\rho)\), uniformly in \(T\). If \(\rho=1\), it grows linearly, and if \(\rho>1\), it can grow geometrically. The architecture alone does not verify a contractive update.

\paragraph{Scope of step-wise supervision.}
The step-wise contrastive objective directly evaluates intermediate retrieval scores, whereas the final-only objective directly evaluates only the terminal readout. This distinction motivates intermediate supervision, but does not prove it is necessary for early exit. Nor does minimizing contrastive loss alone establish a prescribed distance to an ideal intermediate embedding. Both shared parameters and other auxiliary objectives can influence intermediate states without a step-wise contrastive term.

\paragraph{Proposition 2: ranking preservation under an embedding-distance bound.}
For a fixed normalized query \(q\), let \(p\) be the positive candidate and \(n\) index negatives. Define
\begin{equation}
\gamma_t
=
q^\top z_t^p-\max_n q^\top z_t^n.
\end{equation}
Suppose \(\gamma_T\ge m>0\) and every candidate satisfies
\(\|z_t-z_T\|_2\le\epsilon\).
If \(2\epsilon<m\), the step-\(t\) representation preserves the correct full-step top-1 decision.

\paragraph{Proof.}
Cauchy--Schwarz gives
\(\left|q^\top(z_t-z_T)\right|\le\epsilon\).
The positive score can decrease by at most \(\epsilon\), and the largest negative score can increase by at most \(\epsilon\). Hence
\begin{equation}
\gamma_t\ge\gamma_T-2\epsilon\ge m-2\epsilon>0.
\end{equation}
\(\square\)

For normalized embeddings,
\(\|z_t-z_T\|_2^2=2(1-z_t^\top z_T)\), linking the distillation objective to embedding distance. However, a small average distillation loss does not guarantee the uniform bound required for every candidate. The proposition assumes a fixed query and does not directly cover a simultaneously changing query-side embedding.

\paragraph{Assumption 2: calibration conditional on the stopping history.}
Let \(\mathcal F_t\) contain the information available to the stopping rule through step \(t\), including the query, observed states, and previous gate outputs. Let \(A_t\) denote agreement between the step-\(t\) and full-step retrieval decisions. Assume the gate output \(p_t\) is measurable with respect to \(\mathcal F_t\) and satisfies
\begin{equation}
\left|p_t-\Pr(A_t\mid\mathcal F_t)\right|
\le\epsilon_g
\end{equation}
almost surely at every eligible step, with \(\epsilon_g\ge0\) and \(0\le\eta\le1\). This history-conditional assumption is stronger than a small average calibration error.

\paragraph{Proposition 3: disagreement risk at a threshold-based stopping time.}
Let \(\widehat T\) be the first step \(t\ge t_{\min}\) with \(p_t\ge\eta\), or \(T\) if no earlier exit occurs. Under Assumption 2,
\begin{equation}
\Pr(A_{\widehat T}^{\,c})
\le
\min\{1,\,1-\eta+\epsilon_g\}.
\end{equation}

\paragraph{Proof.}
For \(t<T\), the event \(\{\widehat T=t\}\) belongs to \(\mathcal F_t\). On this event, calibration and the threshold imply
\(\Pr(A_t^c\mid\mathcal F_t)\le1-\eta+\epsilon_g\).
Since \(A_T\) holds by definition,
\begin{align}
\Pr(A_{\widehat T}^{\,c})
&=
\sum_{t<T}\mathbb E\!\left[
\mathbf 1_{\{\widehat T=t\}}
\Pr(A_t^c\mid\mathcal F_t)
\right] \\
&\le
(1-\eta+\epsilon_g)\Pr(\widehat T<T)
\le
1-\eta+\epsilon_g.
\end{align}
A probability is also at most one. \(\square\)

This is an agreement bound, not a guarantee of correctness against ground truth. It is conditional on Assumption 2, which is not verified by the reported experiments. A larger threshold delays or prevents exit for a fixed gate-output trajectory, but does not by itself prove improved task accuracy.

\paragraph{Storage cost of index-only annotations.}
Suppose a video has \(R\) query annotations, each selecting \(K\) frames of size \(H\times W\) with \(c\) channels. If each raw pixel-channel value occupies \(s\) bytes and each index occupies \(b\) bytes, uncompressed frame copies require \(RKHWcs\) bytes. Index lists require \(RKb\) bytes, giving an additional-annotation storage ratio of
\begin{equation}
\frac{HWcs}{b}.
\end{equation}
Both approaches retain the original video. Compressed image formats, metadata, and caching alter the practical ratio; this comparison concerns duplicated annotation storage, not total dataset size or decoding latency.

\subsection{Limitations}

QASLR depends on frame selection: its latent memory cannot recover evidence that is absent from the selected observations. The query-aware pathway produces query-conditioned representations for bounded candidate scoring. Reusable corpus indexing instead requires query-independent selection and encoding.

The comparisons with sub-batched single-round training show consistent retrieval gains but task-dependent QA gains. Test-time order and memory perturbations measure sensitivity to altered inputs and state updates, rather than identifying the internal reasoning process. Transfer is evaluated within the Qwen family.

Warm-cache latency depends on selection, decoding, candidate count, and hardware. Fixed memory size bounds the persistent state, but does not imply a proportional reduction in total computation. The seed-averaged scores describe empirical differences without establishing statistical significance. Early-exit bounds concern agreement with full-step predictions under the stated assumptions, rather than unconditional ground-truth accuracy.

\end{document}